\documentclass[journal]{IEEEtran}
\usepackage{multirow}
\usepackage{ulem}

\usepackage{tikz}
\usetikzlibrary{decorations.pathreplacing}
\usetikzlibrary{fadings}

\usepackage{graphicx}

\begin{document}
%
\title{Using Deep Learning Techniques and Inferential Speech Statistics for AI Synthesised Speech Recognition\\}
%
%
%

\author{Arun~Kumar~Singh,~\IEEEmembership{Student Member,~IEEE,}
        and~Priyanka~Singh,~\IEEEmembership{Member,~IEEE}
        and~Karan~Nathwani,~\IEEEmembership{Member,~IEEE}
}

\maketitle

\begin{abstract}
The recent developments in technology has rewarded us with amazing audio synthesis models like TACOTRON and WAVENETS. On the other side, it poses greater threats such as speech clones and deep fakes, that may go undetected. To tackle these alarming situations, there is an urgent need to propose models that can help discriminate a synthesized speech from an actual human speech  and also identify the source of such a synthesis. Here, we propose a model based on Convolutional Neural Network (CNN) and Bidirectional Recurrent Neural Network (BiRNN) that helps to achieve both the aforementioned objectives. The temporal dependencies present in AI synthesised speech are exploited using Bidirectional RNN and CNN. The model outperforms the state-of-the-art approaches by classifying the AI synthesised audio from real human speech with error rate of $\simeq$ 1.9\% and detecting the underlying architecture with an accuracy of $\simeq$ 97\%.

\end{abstract}


%
\IEEEpeerreviewmaketitle

\section{Introduction}
%
%
%
%
\IEEEPARstart{R}{ecent} advancements in the field of AI has generated very realistic and natural type AI synthesised speech and audio \cite{b2,b4}. Most of the synthesised speeches are generated using powerful AI algorithms and training of deep neural networks. There are so many cloned speeches and dangerous deep fakes flooding everywhere, that it causes an urgent concern for authenticating the digital data prior to putting trust in it's content. Though the research in speech forensics has expedited in the last decade, still the literature presents limited research that deals with synthetic speech generated using well known applications like Baidu's text to speech, Amazon's Alexa, Google's wave-net, Apple's Siri, etc. \cite{b7}. The speech generation methods using deep neural nets has become so common that free open source code are readily available for generation of synthetic audios. Many small startups and developers has come up with improved versions of these technology that are producing realistic human like speeches. Major synthetic speech detection works have focused on famous text to speech (TTS) systems. The other not too famous methods have gone unnoticed that have a potential to produce considerable good quality of synthetic speech. 

Few schemes in the literature  demonstrated speech spoofing \cite{b6} and tampering, but not precisely  detecting AI synthesized speech. Hany illustrated in his work how tampering of a digital signal induces correlations in the higher-order statistics but didn't discuss about AI synthesized content \cite{b1}.  Researcher at Google proposed a WaveNet Architecture \cite{b13} to generate synthetic speech that completely revolutionised the speech synthesis using text. Nowadays, most of the devices rely on speech applications for authentication that raises more  concern for security. It is not just sufficient to detect the synthetic speech but also identify the architecture used to generate a specific synthesized speech.  \\
During synthesis of speech, first-order Fourier coefficients or second-order power spectrum correlations can be easily manipulated to match the human speech.  But third-order bispectrum correlations can help to discriminate between human and AI speech. Comparison of various features for differentiating AI synthesised speech and human speech was presented in \cite{b3}. However, these features were handpicked and present no comparison with advanced deep learning algorithms. Muda et al. presented the distinction between male and female speeches using Mel Frequency Cepstral Coefficient (MFCC) \cite{b8}.  MFCC are useful features to identify vocal tracts. Synthetic speech detection using the temporal modulation technique was presented in \cite{b9}. However, we found that including two primary features related to the MFCC: $\Delta$-Cepstral and $\Delta^{2}$-Cepstral, that previous studies haven't reported, increased the  discrimination accuracy significantly. \

Detection of spoofed speech using hand crafted features and classification based on traditional Gaussian mixture models (GMMs) was proposed in \cite{b14}.  Another scheme presenting hand picked features i.e bicoherence magnitude and phase and testing over few data samples  was done in \cite{b15}. Using automatic feature selection using various deep learning models can avoid the extraneous task of choosing handpicked features. CNNs are one of the fundamental models that are widely used in the field of image processing, face recognition \cite{b17}, classification \cite{b16}, pattern recognition and also, in audio processing applications such as speech recognition \cite{b18} and emotion recognition \cite{b19}. \\
Representing and analysing speech from the spectrogram images is the usual method to interpret their characteristic features and metrics in audio waveform. Exploiting spectrogram for synthetic speech can show the inaccurate modelling of high frequency regions and detailed spectral information. Different AI synthesizers have different artifacts and deficiencies, so using CNNs for automatic feature extraction and classification is the right choice in performing this task. The underlying architecture of AI synthesizers induce long range temporal dependencies in synthetic speeches \cite{b1}, \cite{b21}. Also, bidirectional LSTMs and RNNs are quite good in identifying the dependencies between the features \cite{b20}. Based on these studies, we propose to use BiRNNs fused with CNN for classifying and identifying the AI synthesized speech.

A set of hand picked features to distinguish AI synthesised and human speech was done as an initial work by us \cite{b22}. It was limited to a small dataset of speech samples that we collected from some not so famous, open source synthesizers named Natural Reader, Replica AI, and Spik AI. One major problem in this research domain is lack of standard datasets to test over. One naive extension would be to test these set of features on a large dataset that has enough variability. In this work, our first step was to collect a  sufficiently large dataset. The speech samples we collected has more variations like having different accents, gender, and more noisy samples, all mixed together. The reason for choosing variations and uncleaned data was to develop a more robust technique that can work on real life cases.  Also, these results must be compared with other state-of-the-art features and obtained with advanced deep learning techniques like CNNs and RNNs. 



The proposed work is basically divided into two main parts: 
\begin{itemize}
    \item  Part 1: In this phase, we hand picked features to distinguish AI verses human speech and experimented with machine learning models. 
    \item Part 2: In this phase, we experiment with advanced deep leaning techniques to distinguish AI and human speech and choose the most suitable architecture.
\end{itemize} 
Once we are done with both the phases, comparison of the results are done and the most efficient techniques for  discriminating AI from human speech is discussed in detail.  

For Part 1, we employ machine learning algorithms and combine multiple primary features to account for the enhanced discrimination between human speech and  AI synthesized speech. Here, we have integrated Bispectral analysis and Mel Cepstral analysis. Bispectral Analysis can identify higher-order correlations in AI synthesized speeches that are absent in human speech. These higher-order correlations are present in AI synthesized speech due to the effect of neural network architecture used in the process of synthesising speech. As in the process of AI synthesised speech, different passes from the layers of neural network induces some correlations in the AI synthesized speech which are not present in the recorded human speech. These correlations are hard to remove by some manipulations, as they are induced due to the fundamental properties of the synthetic speech generating process \cite{b1}.

Cepstral analysis can identify components in human speech which are not present in AI synthesized speech. Mel Cepstral analysis of the speech reveals strong power components in human speech, which is not present in the AI synthesized speech. The power components may be present in human speech due to the vocal tract, which in contrast, is not the case with AI-Synthesised speech. We perform both these techniques independently on each sample speech and combine the features to classify the human and AI speech. Higher-order spectral correlation revealed by Bispectral analysis and MFCC,  and  $\Delta$-Cepstral and $\Delta^{2}$-Cepstral obtained by Mel Cepstral analysis serve as our classification parameters. 

For Part 2, we have used deep learning models,  Convolutional Neural Network (CNN) along with Bidirectional Recurrent Neural Network (RNN) on a sufficiently large size of data-set for the identification of AI synthesised speech as well as distinguishing from real human speech. We have considered synthesised samples from some freely available services that are not too famous methods. Also, collected speech samples with various variations. We have achieved significantly  good accuracy in identification which is at par with the current state-of-the-art methods.

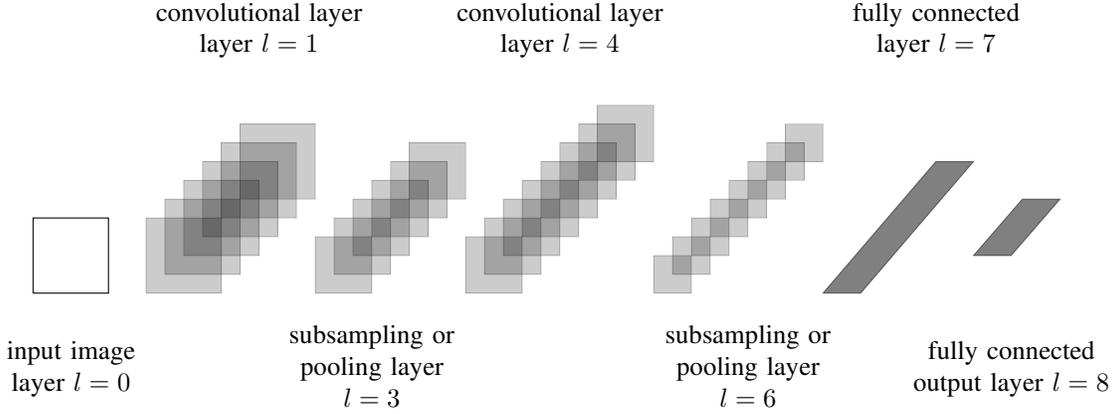
\begin{figure*}[ht]
	\centering
	\begin{tikzpicture}
		\node at (0.5,-1){\begin{tabular}{c}input image\\layer $l = 0$\end{tabular}};
		
		\draw (0,0) -- (1,0) -- (1,1) -- (0,1) -- (0,0);
		
		\node at (3,3.5){\begin{tabular}{c}convolutional layer\\layer $l = 1$\end{tabular}};
		
		\draw[fill=black,opacity=0.2,draw=black] (2.75,1.25) -- (3.75,1.25) -- (3.75,2.25) -- (2.75,2.25) -- (2.75,1.25);
		\draw[fill=black,opacity=0.2,draw=black] (2.5,1) -- (3.5,1) -- (3.5,2) -- (2.5,2) -- (2.5,1);
		\draw[fill=black,opacity=0.2,draw=black] (2.25,0.75) -- (3.25,0.75) -- (3.25,1.75) -- (2.25,1.75) -- (2.25,0.75);
		\draw[fill=black,opacity=0.2,draw=black] (2,0.5) -- (3,0.5) -- (3,1.5) -- (2,1.5) -- (2,0.5);
		\draw[fill=black,opacity=0.2,draw=black] (1.75,0.25) -- (2.75,0.25) -- (2.75,1.25) -- (1.75,1.25) -- (1.75,0.25);
		\draw[fill=black,opacity=0.2,draw=black] (1.5,0) -- (2.5,0) -- (2.5,1) -- (1.5,1) -- (1.5,0);
		
		\node at (4.5,-1){\begin{tabular}{c}subsampling or\\ pooling layer\\ $l = 3$\end{tabular}};
		
		\draw[fill=black,opacity=0.2,draw=black] (5,1.25) -- (5.75,1.25) -- (5.75,2) -- (5,2) -- (5,1.25);
		\draw[fill=black,opacity=0.2,draw=black] (4.75,1) -- (5.5,1) -- (5.5,1.75) -- (4.75,1.75) -- (4.75,1);
		\draw[fill=black,opacity=0.2,draw=black] (4.5,0.75) -- (5.25,0.75) -- (5.25,1.5) -- (4.5,1.5) -- (4.5,0.75);
		\draw[fill=black,opacity=0.2,draw=black] (4.25,0.5) -- (5,0.5) -- (5,1.25) -- (4.25,1.25) -- (4.25,0.5);
		\draw[fill=black,opacity=0.2,draw=black] (4,0.25) -- (4.75,0.25) -- (4.75,1) -- (4,1) -- (4,0.25);
		\draw[fill=black,opacity=0.2,draw=black] (3.75,0) -- (4.5,0) -- (4.5,0.75) -- (3.75,0.75) -- (3.75,0);
		
		\node at (7,3.5){\begin{tabular}{c}convolutional layer\\layer $l = 4$\end{tabular}};
		
		\draw[fill=black,opacity=0.2,draw=black] (7.5,1.75) -- (8.25,1.75) -- (8.25,2.5) -- (7.5,2.5) -- (7.5,1.75);
		\draw[fill=black,opacity=0.2,draw=black] (7.25,1.5) -- (8,1.5) -- (8,2.25) -- (7.25,2.25) -- (7.25,1.5);
		\draw[fill=black,opacity=0.2,draw=black] (7,1.25) -- (7.75,1.25) -- (7.75,2) -- (7,2) -- (7,1.25);
		\draw[fill=black,opacity=0.2,draw=black] (6.75,1) -- (7.5,1) -- (7.5,1.75) -- (6.75,1.75) -- (6.75,1);
		\draw[fill=black,opacity=0.2,draw=black] (6.5,0.75) -- (7.25,0.75) -- (7.25,1.5) -- (6.5,1.5) -- (6.5,0.75);
		\draw[fill=black,opacity=0.2,draw=black] (6.25,0.5) -- (7,0.5) -- (7,1.25) -- (6.25,1.25) -- (6.25,0.5);
		\draw[fill=black,opacity=0.2,draw=black] (6,0.25) -- (6.75,0.25) -- (6.75,1) -- (6,1) -- (6,0.25);
		\draw[fill=black,opacity=0.2,draw=black] (5.75,0) -- (6.5,0) -- (6.5,0.75) -- (5.75,0.75) -- (5.75,0);
		
		\node at (9.5,-1){\begin{tabular}{c}subsampling or\\pooling layer\\ $l = 6$\end{tabular}};
		
		\draw[fill=black,opacity=0.2,draw=black] (10,1.75) -- (10.5,1.75) -- (10.5,2.25) -- (10,2.25) -- (10,1.75);
		\draw[fill=black,opacity=0.2,draw=black] (9.75,1.5) -- (10.25,1.5) -- (10.25,2) -- (9.75,2) -- (9.75,1.5);
		\draw[fill=black,opacity=0.2,draw=black] (9.5,1.25) -- (10,1.25) -- (10,1.75) -- (9.5,1.75) -- (9.5,1.25);
		\draw[fill=black,opacity=0.2,draw=black] (9.25,1) -- (9.75,1) -- (9.75,1.5) -- (9.25,1.5) -- (9.25,1);
		\draw[fill=black,opacity=0.2,draw=black] (9,0.75) -- (9.5,0.75) -- (9.5,1.25) -- (9,1.25) -- (9,0.75);
		\draw[fill=black,opacity=0.2,draw=black] (8.75,0.5) -- (9.25,0.5) -- (9.25,1) -- (8.75,1) -- (8.75,0.5);
		\draw[fill=black,opacity=0.2,draw=black] (8.5,0.25) -- (9,0.25) -- (9,0.75) -- (8.5,0.75) -- (8.5,0.25);
		\draw[fill=black,opacity=0.2,draw=black] (8.25,0) -- (8.75,0) -- (8.75,0.5) -- (8.25,0.5) -- (8.25,0);
		
		\node at (12,3.5){\begin{tabular}{c}fully connected \\layer $l = 7$\end{tabular}};
		
		\draw[fill=black,draw=black,opacity=0.5] (10.5,0) -- (11,0) -- (12.5,1.75) -- (12,1.75) -- (10.5,0);
		
		\node at (13,-1){\begin{tabular}{c}fully connected\\output layer $l = 8$\end{tabular}};
		
		\draw[fill=black,draw=black,opacity=0.5] (12.5,0.5) -- (13,0.5) -- (13.65,1.25) -- (13.15,1.25) -- (12.5,0.5);
	\end{tikzpicture}
	\caption[Architecture of a traditional convolutional neural network.]{The architecture of the original convolutional neural network, as introduced by LeCun}
	\label{fig:traditional-convolutional-network}
\end{figure*}

The remaining paper is structured as follows. Section \ref{sec:Methods} gives a brief overview of the key concepts used in the proposed algorithm, section \ref{sec:Classi} provides the details of the dataset, classification model and parameters used. Section \ref{sec:Results} discusses the result findings of the proposed method.

\section{Preliminaries}\label{sec:Methods} 
In this section, we have given a brief overview of the higher-order statistics that we have used in our proposed algorithm as distinguishing features in experiments of Part 1. Analysis of Mel Frequency Cepstral Coefficient (MFCC) and visualization using the Mel spectrogram is described. Delta and Delta Square related to Mel Cepstrum is briefed. Other than that we have also given a brief introduction to CNN and RNN that we used in experiments of Part 2.

\subsection{Bispectral Analysis}

The simple way to represent the first-order correlation or first-order statistics in Fourier domain is by using Fourier coefficients. 
A Fourier transform for an audio signal $y(k)$ is given by decomposing it into different frequencies:
\begin{equation}
Y(\omega) = \sum_{k=-\infty}^{\infty} y(k).e^{-ik\omega}
\end{equation}
with $\omega\ \epsilon\ [-\pi,\pi]$. The second order correlations are detected generally, by Power spectrum of the signal $P(\omega)$ given by :
\begin{equation}
P(\omega) = Y(\omega).Y^{*}(\omega)
\end{equation}
where $*$ denotes the complex conjugate. But higher order correlations cannot be be detected using Power spectrum because power spectrum is blind to higher order correlations. However, these higher order correlations can be detected using bispectral analysis. We find bispectrum of the signal to calculate third order correlations which is given by:
\begin{equation}
B(\omega_{1},\omega_{2}) = Y(\omega_{1}).Y(\omega_{2}).Y^{*}(\omega_{1} + \omega_{2})
\end{equation}
Unlike the power spectrum, the bispectrum in the Equation (3) is a complex valued quantity. So for the purpose of simplicity and interpretation for our problem, it is suitable to represent or use the complex bispectrum with respect to it's magnitude : 
\begin{equation}
\mid B(\omega_{1},\omega_{2}) \mid\ =\ \mid Y(\omega_{1})\mid .\mid Y(\omega_{2})\mid .\mid Y(\omega_{1} + \omega_{2})\mid
\end{equation}
and Phase :
\begin{equation}
\angle B(\omega_{1},\omega_{2}) = \angle Y(\omega_{1}) + \angle Y(\omega_{2}) - \angle Y(\omega_{1} + \omega_{2})
\end{equation}
Also for the purpose of scaling and simplicity in calculations, it will be helpful to use the normalized bispectrum \cite{b5}, the bicoherence : 
\begin{equation}
B_{c}(\omega_{1},\omega_{2}) = \frac{Y(\omega_{1}).Y(\omega_{2}).Y^{*}(\omega_{1} + \omega_{2})}{\sqrt{\mid Y(\omega_{1}).Y(\omega_{2})\mid^{2}.\mid Y(\omega_{1} + \omega_{2})\mid^{2}}}
\end{equation}
This normalized bispectrum yields magnitude in the range $[0,1]$. But we have used the other normalized process for bispectral magnitude and phase which also yields the range into $[0,1]$. \ 

For the process of normalization, we divide each speech samples of length $N$ into approximately $K \approx 100$ smaller samples of length $N/K$. The bispectral magnitude and phase of these $K$ segments are summed over different $\omega$ values and average value is taken. 
\begin{equation}
 \mid \widehat B(\omega_{1},\omega_{2}) \mid\ =\ \frac{1}{K} \sum_{K}( \mid Y_{K}(\omega_{1})\mid\mid Y_{K}(\omega_{2})\mid\mid Y_{K}(\omega_{1} + \omega_{2})\mid) 
\end{equation}
\begin{equation}
\angle \widehat B(\omega_{1},\omega_{2}) =  \frac{1}{K} \sum_{K}(\angle Y_{K}(\omega_{1}) + \angle Y_{K}(\omega_{2}) - \angle Y_{K}(\omega_{1} + \omega_{2})) 
\end{equation}
After obtaining average values of bispectral magnitude and phase we find the maximum $(max)$ and minimum $(min)$ of all averages for magnitude and phase values respectively. Then for each value of magnitude and phase we subtracted $min$ value from them and divide them with $max$ value resulting into all values normalised to the scale $[0,1]$

\begin{figure*}[ht!]
  \centering
  \includegraphics[scale=0.8]{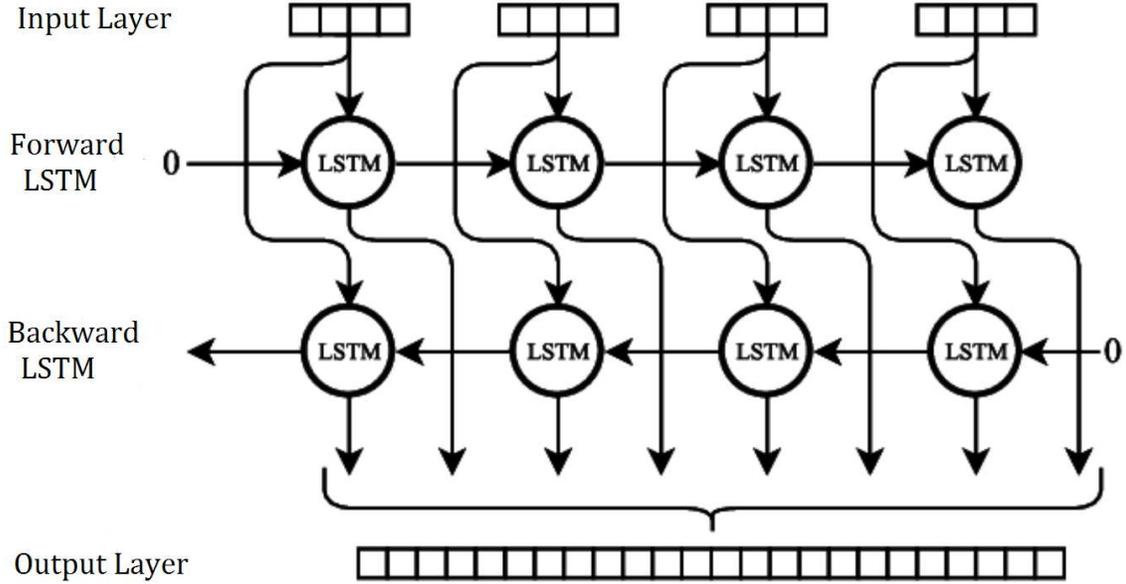}
  \caption{A traditional diagram represnting a BiLSTM architecture showing both forward and backward LSTM.}
  \label{fig:traditional-BiLSTM}
\end{figure*}

\subsection{Mel Frequency Cepstral Coefficient (MFCC) and Analysis}

The speech generated by humans can be uniquely identified due to the vocal tract's shape that includes human oral organs, during the speaking. In general, these different vocal tract shapes helps in determining the type of sound that we speak. The Mel Frequency Cepstral represents the short-time power spectrum of the audio. It can be used to filter speech based on vocal tract shape. This is represented using MFCCs. On a similar hypothesis, there will be differences in MFCCs values of Human speech and AI-synthesized speech as the speech generated by AI are not generated from vocal tracts.

 For our study,  we have considered four types of speech. The first one is human speech, and the other three are AI synthesized speech from three different sources, namely Spik.AI, Natural Reader, and Replica AI. These three kinds of AI speech are generated from different text to speech-generating AI engines.  Mel spectrogram for the four types of speeches is represented in Figure \ref{fig2}.
\  

The MFCCs are calculated from the magnitude spectrum of short term Fourier transform of the audio signal. The short term Fourier transform of the audio signal $y(k)$ is given by :
\begin{equation}
Y(\omega) =\ \mid Y(\omega) \mid e^{\ j\phi(\omega)}
\end{equation}
where $\mid Y(\omega) \mid$ is magnitude spectrum and $\phi(\omega)$ is the phase spectrum. For calculating MFCC, the entire speech is split into overlapping segments called windows. After that, a Fourier transform is performed for each segment, which is used to derive the power spectrum. Mel Frequency Filter is applied to the power spectrum obtained, and then discrete cosine transform (DCT) of the Mel log power is taken. The MFCCs represents the amplitude of the obtained spectrum after DCT is performed.\

Other parameters associated with MFCC useful as a feature for the distinction of speech are $\Delta$-Cepstrum and $\Delta^{2}$-Cepstrum. Change in MFCC coefficients is given by $\Delta$-Cepstrum, i.e., $\Delta$-Cepstrum is the difference between the current MFCC coefficient and the previous MFCC coefficient. Similarly, Change in $\Delta$-Cepstrum values is given by $\Delta^{2}$-Cepstrum, i.e., $\Delta^{2}$-Cepstrum is the difference between current $\Delta$-Cepstrum value and the previous $\Delta$-Cepstrum value. All these three parameters act as strong 
traits for representing Cepstral Analysis.

\subsection{Convolutional Neural Networks (CNN)}

The CNNs are the neural networks that uses linear matrix operation called Convolution to extract meaningful features from the image. The underlying model learns the inter-pixel relations by identifying different patterns in the image. The higher level understanding of CNN architecture says that initial layers divides the input image into smaller segments and then convolution operation uses them as a learnable filters. These forms the convolutional layers. Convolutional layers when combined with normal dense layers forms the strong CNN architectures used to identify more complex patterns, shapes, images and objects.

The CNN usually takes an image with three dimensions as a Input i.e with X rows, Y columns and 3 channels (Red, Green and Blue). The input image goes sequentially for the series of processing. Each processing unit or step is represented using layers. These layers may be Resizing Layer, Normalisation Layer, Convolutional layer or Pooling Layer, etc. The basic architecture of the CNN can be represented by a below equation.
\begin{equation}
x^1 \rightarrow \fbox{$w^1$} \rightarrow x^2 \rightarrow \fbox{$w^2$} \rightarrow \cdots \fbox{$w^{N-1}$} \rightarrow x^N \rightarrow \fbox{$w^N$} \rightarrow z
\end{equation}
In above equation $x$ represents the input that is usually an image and each square box represents the different layers of CNN. The corresponding collective weight for each layer is represented with $w$ inside the square box. The initial input $x^1$ goes to the first layer $w^1$ and output from the first layer $x^2$ act as the input to the second layer processing. This sequence goes on until the last layer $w^N$ and finally the output is given as $z$. The last layer is usually the classification layer. For the image classification problem having C different classes the CNN last layer will have C classes to represent the output. A high level illustration of CNN be seen in Figure \ref{fig:traditional-convolutional-network} indicating convolutional layers, pooling layers and fully-connected layers without details (number of channels or neurons per layer or the input image size).

\subsection{Recurrent Neural Networks (RNN) and BiLSTM}

The RNNs are the Neural Networks that uses the concept of memory. Recurrent Neural Networks are widely used in time-series forecasting to identify data correlations and patterns due to their property of being able to model short term dependencies. 

The theory behind the RNN models are that the output from one layer will be the input to the same layer itself. What this mechanism does is it provides the architecture a “memory” and we were able to comprehend correlation in sequences. This helps the architecture to remember the past and its present decisions which are influenced by learnings from past experiences. A modified version of RNN, which uses memory cells instead of using classic neurons, called as Long Short-Term Memory (LSTM), composed by several components and processes to provide long term memory. Long Short Term Memory (LSTM) first introduced by Hochreiter and Schmidhuber \cite{b24} to learn long-term dependencies in data sequences. 

The mechanism of output from layer as input to itself helps the RNN to make decisions based on past choices it has made in the same state, this induces the memory state. And these states can be changed accordingly. Long Short-Term Memory (LSTM) is the modified version of RNN; the states could be changed in LSTM so that at given times the memory of states would be more or less. Since the RNN is a simpler system, the intuition gained by analyzing the RNN applies to the LSTM network as well. Importantly, the canonical RNN equations, which we derive from differential equations \cite{b23}, serve as the starting model that stipulates a perspicuous logical path toward ultimately arriving at the LSTM system architecture.

When a deep learning architecture is equipped with a LSTM combined with a CNN, it is typically considered as “deep in space” and “deep in time” respectively, which can be seen as two distinct system modalities. CNNs have achieved massive success in visual recognition tasks, while LSTMs are widely used for long sequence processing problems. Because of the inherent properties (rich visual description, long-term temporal memory and end-to end training) of a convolutional LSTM architecture, it has been thoroughly studied for other computer vision tasks involving sequences (e.g. activity recognition or human re-identification in videos) and has lead to significant improvements.

But that all was Unidirectional LSTM. Unidirectional LSTM only preserves information of the past because the only inputs it has seen are from the past. Using bidirectional LSTM will run your inputs in two ways, one from past to future and one from future to past and what differs this approach from unidirectional is that in the LSTM that runs backwards you preserve information from the future and using the two hidden states combined you are able in any point in time to preserve information from both past and future. A sample architecture for BiLSTM can be seen in Figure \ref{fig:traditional-BiLSTM}. We added BiLSTM layers with Convolutional layers to form our model and did our experiments to have a greater efficiency over all other present traditional techniques for our problem. In this paper we have named our model as Convolutional Recurrent Neural Network 32 (CRNN32) as a reference. The number 32 is added as suffix to represent 3 convolutional and 2 BiLSTM layers in our model.

\begin{figure*}[ht!]
  \centering
  \framebox[7in]{\includegraphics[scale=0.65]{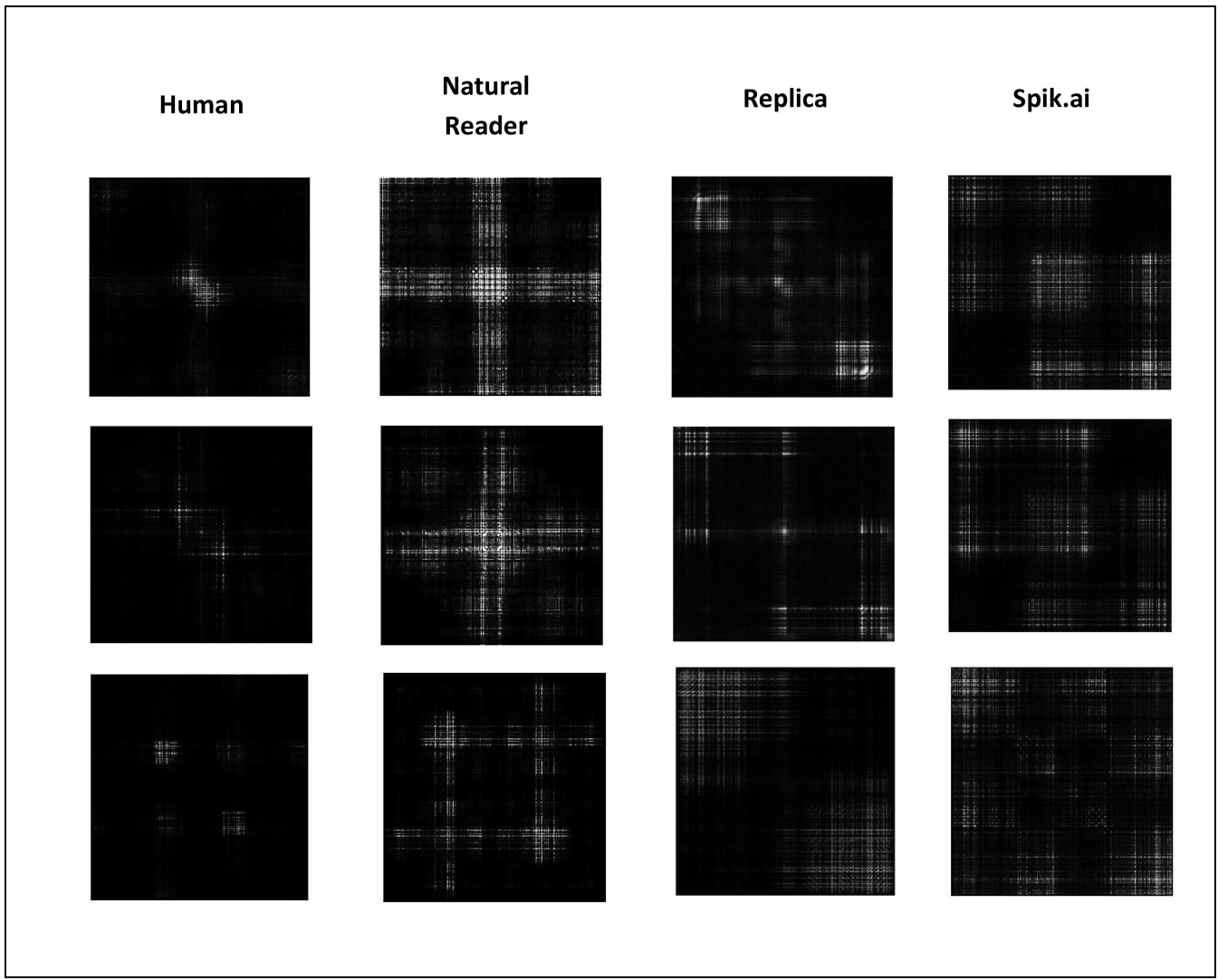}}
  \caption{Bicoherent magnitude of three speakers for human and three synthesized speech. The magnitude plot is shown on an intensity of the scale $[0,1]$}
  \label{fig1}
\end{figure*}

\begin{figure}[ht!]
\centering
\setlength\fboxsep{0.1pt}
\setlength\fboxrule{0.15pt}
\framebox[3.5in]{\includegraphics[scale=0.86]{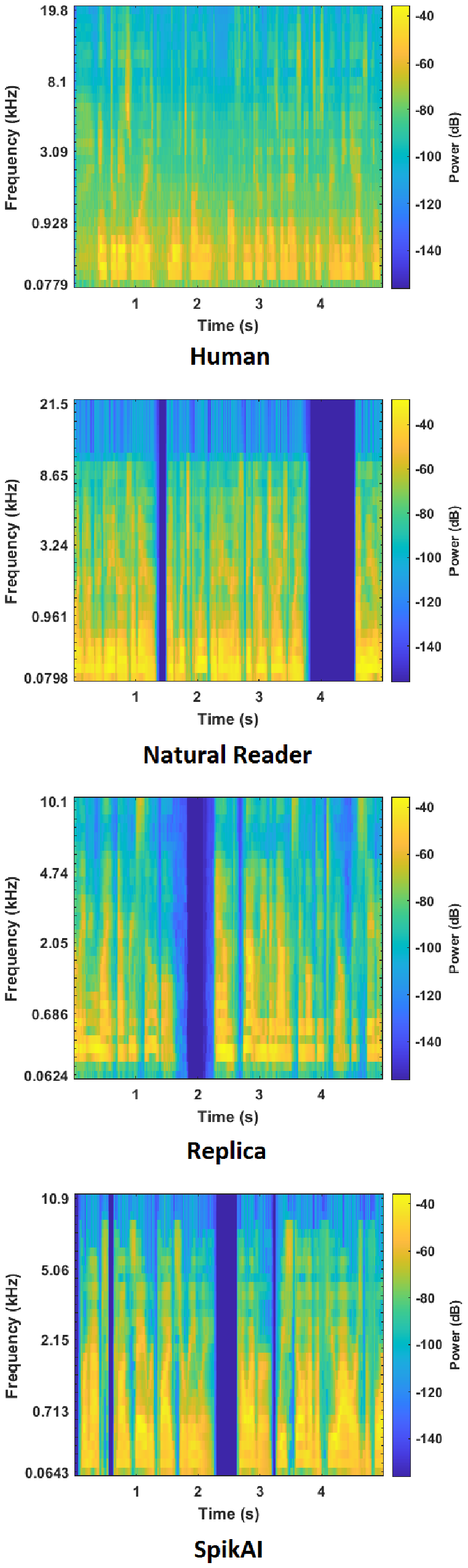}}
\caption{Melspectrogram for different four types of speeches}
\label{fig2}
\end{figure}


\section{Experiments and Analysis}
\label{sec:Classi} 

In this section, we have described the dataset created by us by collecting speech samples from different sources. Some relics for the observations, the classification model, and the parameters used in the study are also described.

\subsection{Dataset}
To perform the experiments, we created a dataset of total 9999 speech samples by collecting it from different sources. For human speech, we have a total of 4862 samples, which consist of speech samples taken from the Mozilla's data repository along with speech samples that are recorded with the microphone. For diversity in our data set, we took human speech samples from both male and female speeches. For AI synthesized speech, we have collected samples from three different sources: Natural Reader, Spik.AI, and Replica. We have taken 3172 samples from Natural Reader,  1543 samples from Spik.AI, and 372 samples from Replica AI. These AI synthesized speech engines are based on text-to-speech synthesis.
\ 
Before processing each speech sample (both AI synthesized and human), it is trimmed into the slots ranging from 4 to 5 seconds. Since audios exist in two channels, i.e., mono and stereo, we have first converted all the speech samples into monotype for uniform comparison. For supervised learning, we took equal samples for each of the four classes i.e. Human, Natural Reader, Replica  and Spik.AI, for training. We took 150 samples for each class, that contributed to a total of 600 samples for training and rest 9399 for testing.

\subsection{Relics for the Observations}
We demonstrated some of the preliminary results in the form of images in Figure \ref{fig1}, showing different patterns that were obtained after processing different speeches. The bicoherence magnitude for three speakers of each category is represented in Figure \ref{fig1}. The first column represents the relic of human speech. Similarly, second, third, and fourth row represent relics from AI synthesized speech, i.e., Spik.AI, Natural Reader, and Replica AI. Similarly, the three different rows represent the normalized bispectral magnitude for three different speakers correspondingly. All images are shown on the same intensity scale.\

On observing the relics, we can see a glaring difference between the magnitude of human speech and all other AI synthesized speech.  These variations can be due to significant spectral correlations present in AI synthesized speech but absent in human speech. These spectral correlations in  AI synthesized speech are induced due to neural network architecture. In particular, the long-range temporal connections between the layers of neural networks may be the cause.\ 

Four Mel spectrograms of different four speeches, i.e., Human, Natural Reader, Replica and SpikAI speech is shown in  Figure \ref{fig2}.  Mel power scale is given on the right of each spectrogram. From reference scale, we can see that dark blue color indicate weak power component and yellow color represents strong power component in the spectrogram \ref{fig2}. It is observed that a strong power component is missing in all types of AI-synthesized speech, which is not in the case of human speech. It may be due to the absence of vocal tract during the generation of  AI-synthesized speech. These differences shown in the Mel spectrogram encourage us to use MFCC as a feature for discriminating speeches.

\subsection{Machine Learning Classification Models and Experiment}

The bicoherence magnitude and phase are calculated, as mentioned in Equation 7 and Equation 8, respectively, for all human speech samples and three types of AI synthesized speech. These quantities are then normalized to the range [0,1] and then used to calculate the machine learning model's higher-order statistical parameters. 

We calculate the first four statistical moments for both magnitude and phase. Let M and P be the random variables denoting the underlying distribution of bicoherence magnitude and phase. The first four statistical moments are given by:\\
\begin{itemize}
    \item Mean , $\mu_{X} = E_{X}[X]$ \\
    \item Variance , $\sigma_{X} = E_{X}[(X-\mu_{X})^{2}]$\\
    \item Skewness , $\gamma_{X} = E_{X}[(\frac{X-\mu_{X}}{\sigma_{X}})^{3}]$\\
    \item Kurtosis , $\kappa_{X} = E_{X}[(\frac{X-\mu_{X}}{\sigma_{X}})^{4}]$\\
\end{itemize}
where $E_{X}[.]$ is the expected value operator with regards to random variable X. For the magnitude, we represent X = M, and for the phase X = P, these four moments are calculated by replacing this expected value operator with average. Also, for each speech sample, the mean and variance of MFFC, $\Delta$-Cepstrum, and $\Delta^{2}$-Cepstrum are calculated. It contributes to a 15-D feature vector for each speech sample. The first 8 entries represent the four moments for magnitude and phase. The next 6 entries represent the mean and variance of MFCC, $ \Delta $-Cepstrum, and $\Delta^{2}$-Cepstrum and last entry represent the class of speech, i.e., Human, Natural Reader, Spik.AI or Replica.

We perform experiments considering following scenarios:
\begin{itemize}
    \item  Scenario 1: In this setup, we classify speech samples into two classes i.e., Human vs. AI-synthesized (Natural Reader, Spik.AI or Replica). It is a binary classification and main focus of this paper.
    \item Scenario 2: In this experiment, we classify speech samples into multiple classes i.e., Human, Natural Reader, Spik.AI and Replica.
\end{itemize}
\
The 15-D feature representation for these experiment scenarios differs only in the last entry. For scenario 1, the last entry can represent either of the two values, i.e., Human or AI synthesized. However, for scenario 2, it can represent any of the four classes, i.e., Human, Natural Reader, Spik.AI, or Replica. We perform machine learning-based classification for both the experiment scenarios with the feature mentioned earlier, with the intuition of different expected outcomes. 



\begin{table*}[ht]
\centering
\begin{tabular}{|c|c|c|c|c|c|c|c|}
\hline
\multicolumn{1}{|l|}{}             & \multicolumn{5}{c|}{\textbf{Individual Features}}                           & \multicolumn{2}{c|}{\textbf{Combined Features}} \\ \hline
\textbf{Various Models} &
  {\begin{tabular}[c]{@{}c@{}}Bicoherence \\ Magnitude\end{tabular}} &
  {\begin{tabular}[c]{@{}c@{}}Bicoherence \\ Phase\end{tabular}} &
  {MFCC} &
  {\begin{tabular}[c]{@{}c@{}}Delta \\ Cepstral\end{tabular}} &
  {\begin{tabular}[c]{@{}c@{}}Delta \\ Sqaure Cepstral\end{tabular}} &
  {\begin{tabular}[c]{@{}c@{}}Bicoherence\\ (Magnitude \& Phase)\end{tabular}} &
  {\begin{tabular}[c]{@{}c@{}}MFCC\\ \&\\ Delta Cepstral\\ \&\\ Delta Square\\ Cepstral\end{tabular}} \\ \hline
{Fine Tree}                 & 50            & 28.2          & 68.8          & 66.5          & 51.8        & 47.3                   & 77.5                   \\ \hline
{Linear Discriminiant}      & 46.5          & 30.5          & 50.2          & 54.2          & 45.5        & 46.7                   & 64.8                   \\ \hline
{Quadratic Discriminant}    & 44.5          & 27.5          & 63.7          & 45.8          & 28.3        & 43.2                   & 56.2                   \\ \hline
{Gaussian Naïve Bayes}      & 40.8          & 25.5          & 48.2          & 46.7          & 28.8        & 40.8                   & 52.8                   \\ \hline
{Linear SVM}                & 47.8          & 30.5          & 60.3          & 51.7          & 35.5        & 47.8                   & 64                     \\ \hline
{Quadratic SVM}             & 52            & 29.3          & 52.3          & 18.7          & 28.8        & \textbf{52.7}          & 64.2                   \\ \hline
{Weighted KNN}              & \textbf{53.2} & 32.3          & 69.3          & 64.5          & 48.8        & 48.7                   & 75.3                   \\ \hline
{Boosted Trees Ensemble}    & 47.8          & \textbf{36.7} & 70.8          & \textbf{67.2} & \textbf{54} & 49.3                   & \textbf{79.5}          \\ \hline
{Bagged Trees Ensemble}     & 49.8          & 30.8          & \textbf{71.3} & 67            & 53.2        & 51.8                   & 78.8                   \\ \hline
{RUSBoosted Trees Ensemble} & 48.7          & 26.7          & 67.5          & 66.3          & 53.3        & 48.3                   & 74.7                   \\ \hline
\end{tabular} \vspace{2mm}
\caption{MultiClass : Various accuracy of individual features and few of the combined features over training data for different models.}
\label{tab:MultiClass Experiment1}
\end{table*}

\begin{table*}[ht]
\centering
\begin{tabular}{|c|c|c|c|c|c|c|c|}
\hline
\multicolumn{1}{|l|}{}             & \multicolumn{5}{c|}{\textbf{Individual Features}}                             & \multicolumn{2}{c|}{\textbf{Combined Features}} \\ \hline
\textbf{Various Models} &
  {\begin{tabular}[c]{@{}c@{}}Bicoherence \\ Magnitude\end{tabular}} &
  {\begin{tabular}[c]{@{}c@{}}Bicoherence \\ Phase\end{tabular}} &
  {MFCC} &
  {\begin{tabular}[c]{@{}c@{}}Delta \\ Cepstral\end{tabular}} &
  {\begin{tabular}[c]{@{}c@{}}Delta \\ Sqaure Cepstral\end{tabular}} &
  {\begin{tabular}[c]{@{}c@{}}Bicoherence\\ (Magnitude \& Phase)\end{tabular}} &
  {\begin{tabular}[c]{@{}c@{}}MFCC\\ \&\\ Delta Cepstral\\ \&\\ Delta Square\\ Cepstral\end{tabular}} \\ \hline
{Fine Tree}                 & 70.8          & 65            & 83            & 91            & 85.3          & 71.2                  & 90.5                    \\ \hline
{Linear Discriminiant}      & 74.5          & 75            & 76.7          & 75            & 75            & 75.2                  & 76.3                    \\ \hline
{Quadratic Discriminant}    & 77.2          & 73.3          & 61.5          & 74.2          & 33.2          & 75.8                  & 62.7                    \\ \hline
{Gaussian Naïve Bayes}      & 75.3          & 73.8          & 59.7          & 74.2          & 33.2          & 75.3                  & 58.8                    \\ \hline
{Linear SVM}                & 75.2          & 75            & 73.8          & 66.3          & 60.2          & 75.2                  & 76.3                    \\ \hline
{Quadratic SVM}             & 75.3          & 75            & 60.8          & 60.5          & 53            & 76.5                  & 64.7                    \\ \hline
{Weighted KNN}              & \textbf{77.3} & 75.3          & 80.2          & 86.7          & 79.3          & 76.2                  & 90.2                    \\ \hline
{Boosted Trees Ensemble}    & 75.8          & 74.2          & 87            & 91.2          & 82.8          & \textbf{78}           & \textbf{93.5}           \\ \hline
{Bagged Trees Ensemble}     & 76.7          & \textbf{75.7} & \textbf{86.2} & \textbf{91.8} & \textbf{83.3} & 75                    & 91.3                    \\ \hline
{RUSBoosted Trees Ensemble} & 72.2          & 56.3          & 82.3          & 90.5          & 81.7          & 72                    & 92.2                    \\ \hline
{Logistic Regression}       & 75            & 75            & 77.7          & 74            & 74.2          & 74.8                  & 77                      \\ \hline
\end{tabular} \vspace{2mm}
\caption{BinaryClass : Various accuracy of individual features and few of the combined features over training data for different models.}
\label{tab:BinaryClass Experiment1}
\end{table*}

By visualizing the data for both types of classification, we tried a few of the learning algorithms to train the model for collecting data based on our intuition. For both binary class and multi-class classifications, we did several experiments to have an insight for better comparison. We majorly categorize our experiments in four sub-tasks according to the feature we took for our experiment. 
\begin{itemize}
    \item  Sub-Task 1: In this task, we considered five features i.e Bicoherence Magnitude, Bicoherence Phase, MFCC, Delta Cepstral and Delta Square Cespstral. We tested the accuracies of these features over different ML algorithms when taken individually which are listed in  Table 1 to Table 5.
    \item Sub-Task 2: In this task, we combined two of the features i.e Bicoherence Magnitude, and Bicoherence Phase and tested their accuracies over different ML algorithms. These were the same features as considered by \cite{b7} in their experiments.  Results are listed down in Table 1 to Table 5.
    \item Sub-Task 3: In this task, we combined three of the features i.e MFCC, Delta Cepstral and Delta Square Cepstral and tested their accuracies over different ML algorithms.  Results are listed in Table 1 to Table 5.
    \item Sub-Task 4: In this task, we combined all the five features that we took individually in sub-task 1 i.e Bicoherence Magnitude, Bicoherence Phase, MFCC, Delta Cepstral and Delta Square Cespstral. We tested thier accuracies on all same ML algorithms that we have used for our other different sub-tasks. The result is listed in Table 6.
\end{itemize}

For both the binary and multi-class classification, we experimented with the following  machine learning algorithms:  Linear Discriminant, Quadratic Discriminant, Gaussian Naive Bayes, Quadratic SVM, Linear SVM, Weighted KNN, Boosted Trees, Bagged Trees and RUS-Boosted Trees. These algorithms are used for training, validation and testing. All validations are done using 5-fold cross-validation.

\subsection{Convolutional Recurrent Neural Network 32 (CRNN32) Model and Experiment}


\begin{figure}[ht!]
\centerline{\includegraphics[width=0.52\textwidth]{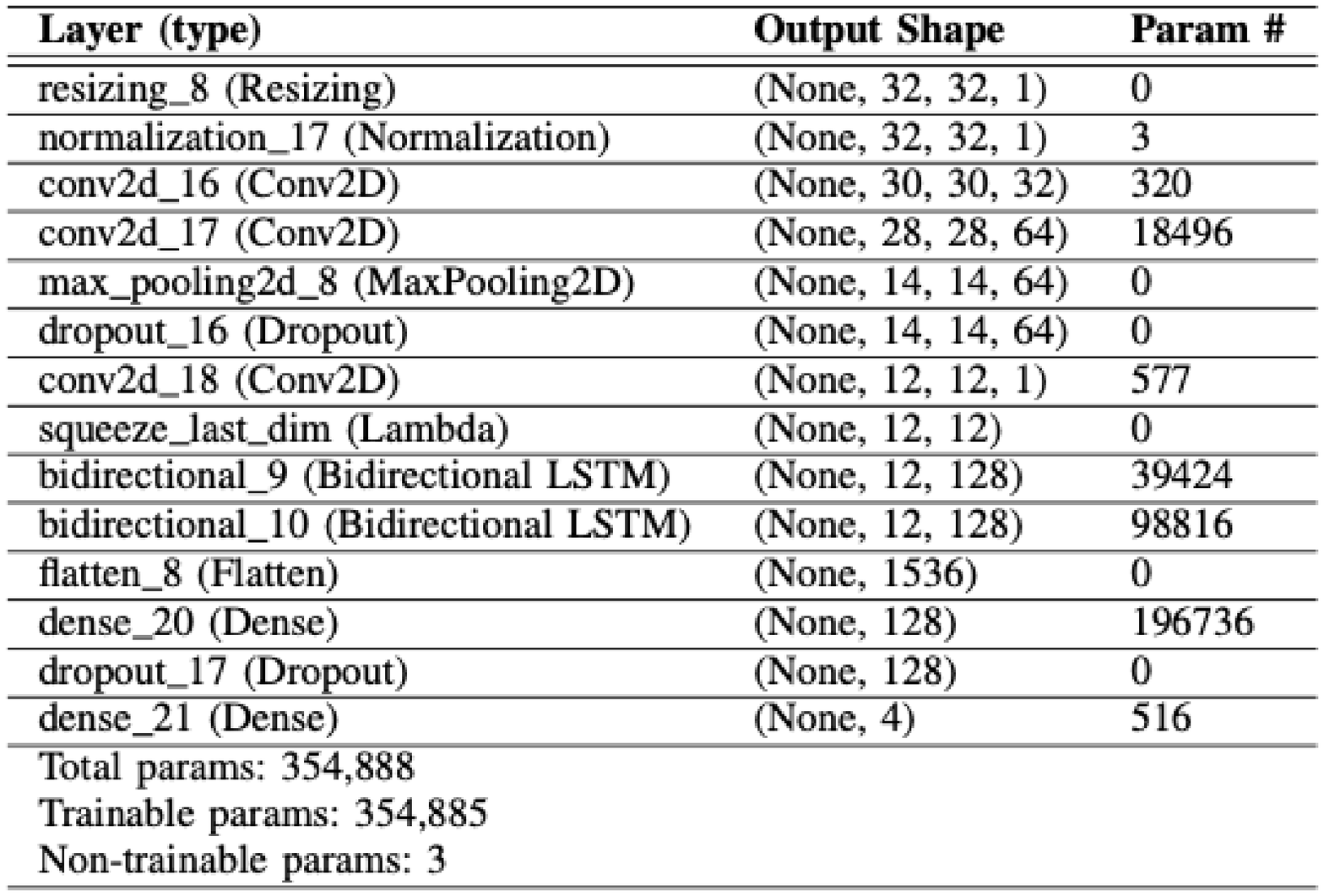}}
\caption{CRNN model summary for multi-class classification}
\label{figCRNN}
\end{figure}

In our proposed model, we fused Convolutional and Bidirectional LSTM layers along with other required layers to create a new model Convolutional Recurrent Neural Network 32 (CRNN32). We have experimented the deep learning architecture for both scenarios, binary class as well as  multi-class classification.  The model summary for the multi class classification is listed in Figure \ref{figCRNN}. For the binary class classification also, the model  used is same as multi-class classification but only difference is in output layer. Binary class model has the same model summary as mentioned in Figure \ref{figCRNN} with last dense layer having the size of 2, because of only two output classes.

The input to the model is the spectrogram image of the speech samples from the data set. We used Kera's sequential models for Tensorflow to build our model. In our model, we added the first layer as the Resizing layer that re-sizes all the input spectrogram image into the size of $32 \times 32 \times 1$ for processing followed by normalization layer which normalizes the features into one range to speed up the learning process. After that, we have added two convolutional layers of size ($30 \times 30 \times 32$) and ($28 \times 28 \times 64$) respectively. These two convolutional layers will try to extract the learnable features from the input samples for the speech and will update the weights for each neuron of the layers correspondingly. Then we have used the Max pooling layer followed by dropout layer before adding our third convolutional layer. The Max pooling layer and dropout layer is added to avoid over fitting and reduce the computational cost by filtering out the required only learnable parameters. The third convolution layer will now, again try to extract fractures from the output after dropout. After the convolutional layer, we have put the lambda layer which squeezes the dimensions of the input suitable to be passed for Bidirectional LSTM layers. Then, we added two Bidirectional LSTM layers having $39424$ and $98816$ learnable parameters respectively. Output received from second Bidirectional LSTM layer is flattened into the one dimension tensor of size $1536$ using flatten layer. After that, we have added a dense layer that is basic simple neural network layer which will try to converge the output to the required features. Again, another dropout layer is added to drop out the unnecessary features to avoid over fitting of the model. At last, we have added a simple dense layer of size $4$ neurons each, corresponding to one of the four classes of the multi- class classification. In binary class, this last dense layer has size $2$ corresponding to two different classes `Human' vs `AI speech'. 

For performing experiments, we divided the dataset into 3 parts, each corresponding to training, validation, and testing. The percentage from total data for these three divisions are 60\%, 20\% and 20\% respectively.  After completing training and validation over our 80\% (60 + 20) of data, we test our trained model over remaining 20\% of test data. 

\section{Results and Discussions}
\label{sec:Results} 

In this section, we have presented the result analysis and discussion for both our machine learning and deep learning based experiments. For both the experiments, the individual analysis of the results are presented in separate subsections.

\begin{table*}[ht]
\centering
\begin{tabular}{|c|c|c|c|c|c|c|c|}
\hline
\multicolumn{1}{|l|}{}             & \multicolumn{5}{c|}{\textbf{Individual Features}}                                 & \multicolumn{2}{c|}{\textbf{Combined Features}} \\ \hline
\textbf{Various Models} &
  {\begin{tabular}[c]{@{}c@{}}Bicoherence \\ Magnitude\end{tabular}} &
  {\begin{tabular}[c]{@{}c@{}}Bicoherence \\ Phase\end{tabular}} &
  {MFCC} &
  {\begin{tabular}[c]{@{}c@{}}Delta \\ Cepstral\end{tabular}} &
  {\begin{tabular}[c]{@{}c@{}}Delta \\ Sqaure Cepstral\end{tabular}} &
  {\begin{tabular}[c]{@{}c@{}}Bicoherence\\ (Magnitude \& Phase)\end{tabular}} &
  {\begin{tabular}[c]{@{}c@{}}MFCC\\ \&\\ Delta Cepstral\\ \&\\ Delta Square\\ Cepstral\end{tabular}} \\ \hline
{Fine Tree}                 & 45.95          & \textbf{28.98} & 65.61         & 64.83          & 56.11          & 48.16                  & 77.79                  \\ \hline
{Linear Discriminiant}      & \textbf{62.31} & 16.01          & \textbf{76.1} & 65.24          & \textbf{66.76} & 57.27                  & 79.48                  \\ \hline
{Quadratic Discriminant}    & 46.05          & 13.68          & 51.56         & 17.09          & 15.93          & 46.64                  & 48.77                  \\ \hline
{Gaussian Naïve Bayes}      & 54.37          & 11.13          & 30.8          & 17.02          & 16.1           & 53.54                  & 46.43                  \\ \hline
{Linear SVM}                & 60.19          & 10.6           & 39.53         & 62.69          & 56.47          & \textbf{58.21}         & 80.17                  \\ \hline
{Quadratic SVM}             & 57.34          & 19.77          & 44.38         & 13.4           & 16.43          & 57.83                  & \textbf{81.14}         \\ \hline
{Weighted KNN}              & 51.82          & 28.27          & 63.16         & 58.99          & 51.84          & 54.11                  & 71.75                  \\ \hline
{Boosted Trees Ensemble}    & 46.3           & 24.87          & 67.46         & \textbf{69.22} & 51.84          & 53.39                  & 75.45                  \\ \hline
{Bagged Trees Ensemble}     & 50.63          & 28.7           & 64.94         & 64.9           & 57.26          & 52.51                  & 78.65                  \\ \hline
{RUSBoosted Trees Ensemble} & 47.52          & 26.11          & 71.89         & 63.7           & 50.55          & 52.16                  & 73.18                  \\ \hline
\end{tabular} \vspace{2mm}
\caption{MultiClass : Various accuracy of individual features and few of the combined features over test data for different models.}
\label{tab:MultiClass Experiment2}
\end{table*}

\begin{table*}[ht]
\begin{tabular}{|c|c|c|c|c|c|c|c|}
\hline
                          & \multicolumn{5}{c|}{\textbf{Individual Features}}                                  & \multicolumn{2}{c|}{\textbf{Combined Features}} \\ \hline
\textbf{Various Models} &
  \begin{tabular}[c]{@{}c@{}}Bicoherence  \\ Magnitude\end{tabular} &
  \begin{tabular}[c]{@{}c@{}}Bicoherence\\ Phase\end{tabular} &
  MFCC &
  \begin{tabular}[c]{@{}c@{}}Delta\\ Cepstral\end{tabular} &
  \begin{tabular}[c]{@{}c@{}}Delta\\ Sqaure Cepstral\end{tabular} &
  \begin{tabular}[c]{@{}c@{}}Bicoherence\\ (Magnitude \& Phase)\end{tabular} &
  \begin{tabular}[c]{@{}c@{}}MFCC\\ \&\\ Delta Cepstral\\ \&\\ Delta Square\\ Cepstral\end{tabular} \\ \hline
Fine Tree                 & 56.83          & 52.97          & 76.19          & \textbf{88.2}  & 79.91          & 62.63                  & 86.99                  \\ \hline
Linear Discriminiant      & 60.83          & 49.33          & 55.13          & 49.32          & 42.29          & 61.21                  & 61.4                   \\ \hline
Quadratic Discriminant    & 51.38          & 49.42          & 86.81          & 49.33          & 59.81          & 51.8                   & 86.77                  \\ \hline
Gaussian Naive Bayes      & 60.59          & 49.34          & 86.67          & 49.33          & 54.82          & 60.67                  & 85.86                  \\ \hline
Linear SVM                & 52.92          & 49.33          & 84.11          & 43.12          & 75.17          & 52.9                   & 90.44                  \\ \hline
Quadratic SVM             & 41.31          & 49.34          & 66.02          & 40.16          & 52.63          & 55.01                  & 43.70                  \\ \hline
Weighted KNN              & 59.41          & 52.55          & 72.86          & 80.55          & 72.55          & 61.14                  & 81.05                  \\ \hline
Boosted Trees Ensemble    & 54.68          & 52.61          & 77.8           & 85.58          & 81.18          & 57.51                  & 87.75                  \\ \hline
Bagged Trees Ensemble     & 56.69          & 52.58          & 76.93          & 85.58          & 78.11          & 57.45                  & 86.24                  \\ \hline
RUSBoosted Trees Ensemble & \textbf{65.77} & \textbf{56.63} & \textbf{86.86} & \textbf{88.12} & \textbf{87.38} & \textbf{69.6}          & \textbf{90.72}         \\ \hline
Logistic Regression       & 59.69          & 49.33          & 68.29          & 49.32          & 49.29          & 60.15                  & 72.12                  \\ \hline
\end{tabular} \vspace{2mm}
\caption{BINARY CLASS: VARIOUS ACCURACY OF INDIVIDUAL FEATURES AND FEW  OF THE COMBINED  FEATURES OVER  TEST DATA  FOR DIFFERENT MODELS.}
\label{tab:BinaryClass Experiment}
\end{table*}

\subsection{Machine Learning Models Based Experiments}

\begin{figure}[ht!]
\centerline{\includegraphics[width=0.52\textwidth]{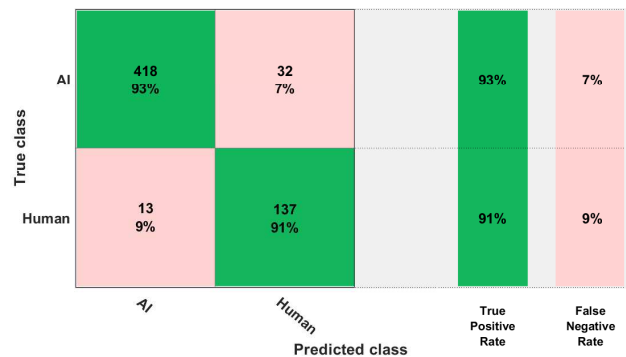}}
\caption{Confusion matrix for binary classification on training data set}
\label{fig6}
\end{figure}

\begin{figure}[ht!]
\centerline{\includegraphics[width=0.52\textwidth]{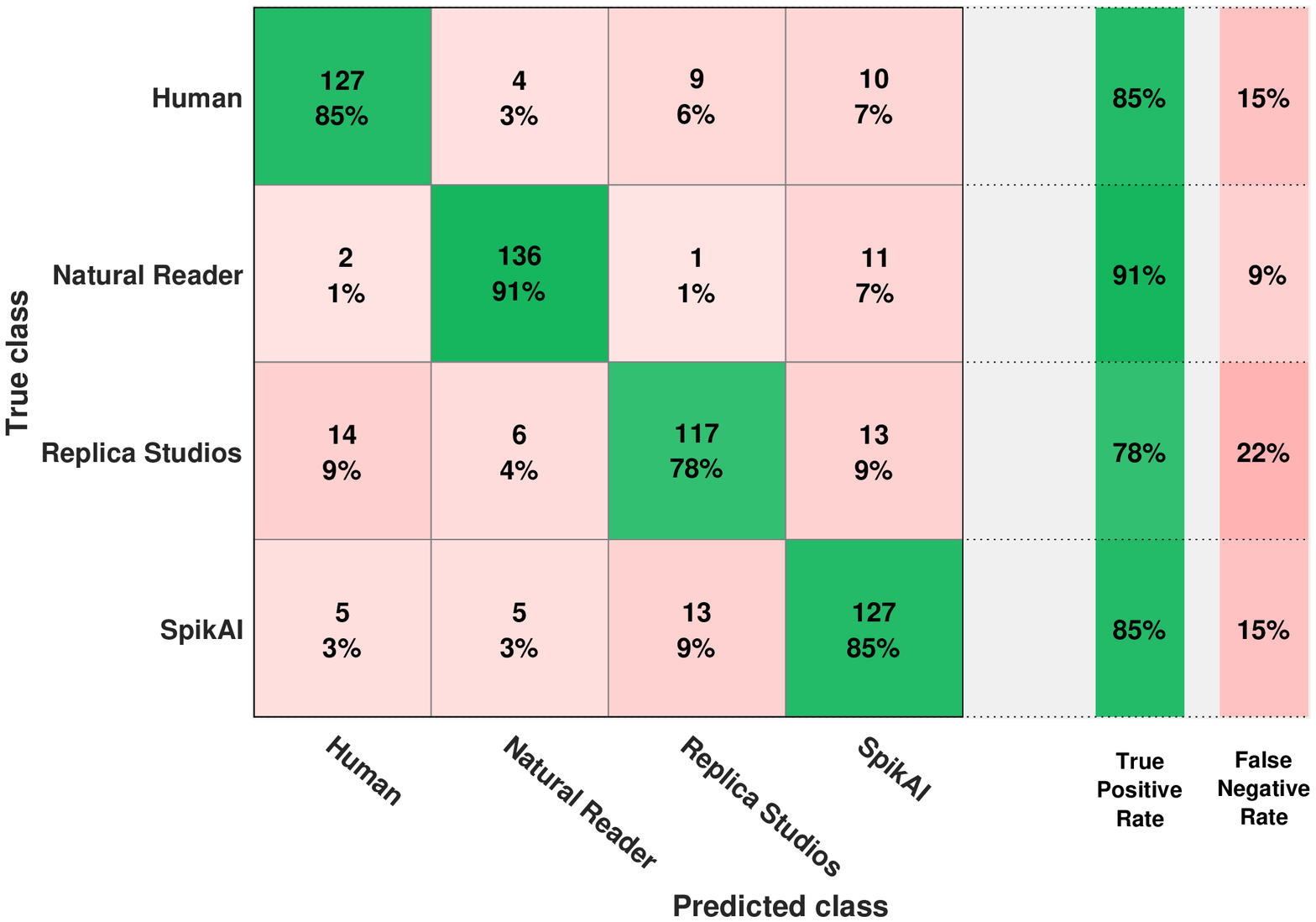}}
\caption{Confusion matrix for multi-class classification on training data set}
\label{fig5}
\end{figure}

After testing each classifier's performance by 5-fold cross-validation, we found that for the binary classification, RUS-Boosted Trees algorithm based machine learning model achieved the highest accuracy of 94.5\%. For multi-class classification, the Bagged Trees algorithm based machine learning model achieved the highest accuracy with 84.5\%. 
Hence, for binary class classification we chose RUS Boosted Trees and for multi-class classification, we chose Bagged Trees as our classifier for constructing the trained ML model. The classifier's exact accuracy can be visualized with a confusion matrix, which indicates how many true values are falsely recognized. The confusion matrix for both the classifier RUS Boosted Trees for binary class and Bagged Tree for multi-class is presented in Figure \ref{fig6} and Figure \ref{fig5}, respectively.

By observing both the confusion matrix, we  can see that binary classification results are more potent than the multi-class classification. In multi-class classification, we have false positives between Natural Reader and Spik.AI, Spik.AI and Replica, etc., which will be treated inside one class, i.e., AI synthesized in binary class classification. In the confusion matrix, we can observe that many of the Natural Reader speech samples are misclassified as Spik.AI samples. This may be due to their similarity in neural network engines. However, this thing does not matter much in the binary classification because all AI synthesized speech is classified as AI synthesized irrespective of what neural network architecture they follow. That is why binary classification between Human speech vs. AI synthesized speech has much better accuracy on cross-validation than multi-class classification. 

\begin{table*}[ht]
\begin{tabular}{|c|c|c|c|c|c|c|c|}
\hline
                          & \multicolumn{5}{c|}{\textbf{Individual Features}}                                       & \multicolumn{2}{c|}{\textbf{Combined Features}} \\ \hline
\textbf{Various Models} &
  \begin{tabular}[c]{@{}c@{}}Bicoherence  \\ Magnitude\end{tabular} &
  \begin{tabular}[c]{@{}c@{}}Bicoherence\\ Phase\end{tabular} &
  MFCC &
  \begin{tabular}[c]{@{}c@{}}Delta\\ Cepstral\end{tabular} &
  \begin{tabular}[c]{@{}c@{}}Delta\\ Sqaure Cepstral\end{tabular} &
  \begin{tabular}[c]{@{}c@{}}Bicoherence\\ (Magnitude \& Phase)\end{tabular} &
  \begin{tabular}[c]{@{}c@{}}MFCC\\ \&\\ Delta Cepstral\\ \&\\ Delta Square\\ Cepstral\end{tabular} \\ \hline
Fine Tree                 & 0.6466          & 0.6272          & 0.7992          & \textbf{0.8922} & 0.8266          & 0.6993                  & 0.8821                \\ \hline
Linear Discriminiant      & \textbf{0.7150} & 0.6607          & 0.6874          & 0.6606          & 0.6603          & 0.7167                  & 0.7183                \\ \hline
Quadratic Discriminant    & 0.6423          & 0.6585          & 0.8585          & 0.6607          & 0.2128          & 0.6427                  & 0.8542                \\ \hline
Gaussian Naive Bayes      & 0.7138          & 0.6607          & 0.8508          & 0.6607          & 0.2131          & 0.7142                  & 0.8424                \\ \hline
Linear SVM                & 0.6770          & 0.6607          & 0.8163          & 0.5978          & 0.6834          & 0.6769                  & 0.9071                \\ \hline
Quadratic SVM             & 0.5653          & 0.6607          & 0.4864          & 0.4479          & 0.1181          & 0.6723                  & 0.5998                \\ \hline
Weighted KNN              & 0.6827          & 0.6592          & 0.7787          & 0.8316          & 0.7774          & 0.7082                  & 0.8359                \\ \hline
Boosted Trees Ensemble    & 0.6633          & \textbf{0.6698} & 0.8117          & 0.8696          & 0.8361          & 0.6718                  & 0.8887                \\ \hline
Bagged Trees Ensemble     & 0.6763          & 0.6635          & 0.806           & 0.8696          & 0.8145          & 0.6806                  & 0.8768                \\ \hline
RUSBoosted Trees Ensemble & \textbf{0.7014} & 0.5819          & \textbf{0.8756} & \textbf{0.8900} & \textbf{0.8815} & \textbf{0.7372}         & \textbf{0.912}        \\ \hline
Logistic Regression       & 0.7093          & 0.6607          & 0.7566          & 0.6606          & 0.6603          & 0.7114                  & 0.7762                \\ \hline
\end{tabular} \vspace{2mm}
\caption{Binary Class: F1 Score for Individual Features and few  of the combined Features over test data for different Machine Learning Models.}
\label{tab:BinaryClass Experiment3}
\end{table*}

\begin{figure}[ht!]
\centerline{\includegraphics[width=0.52\textwidth]{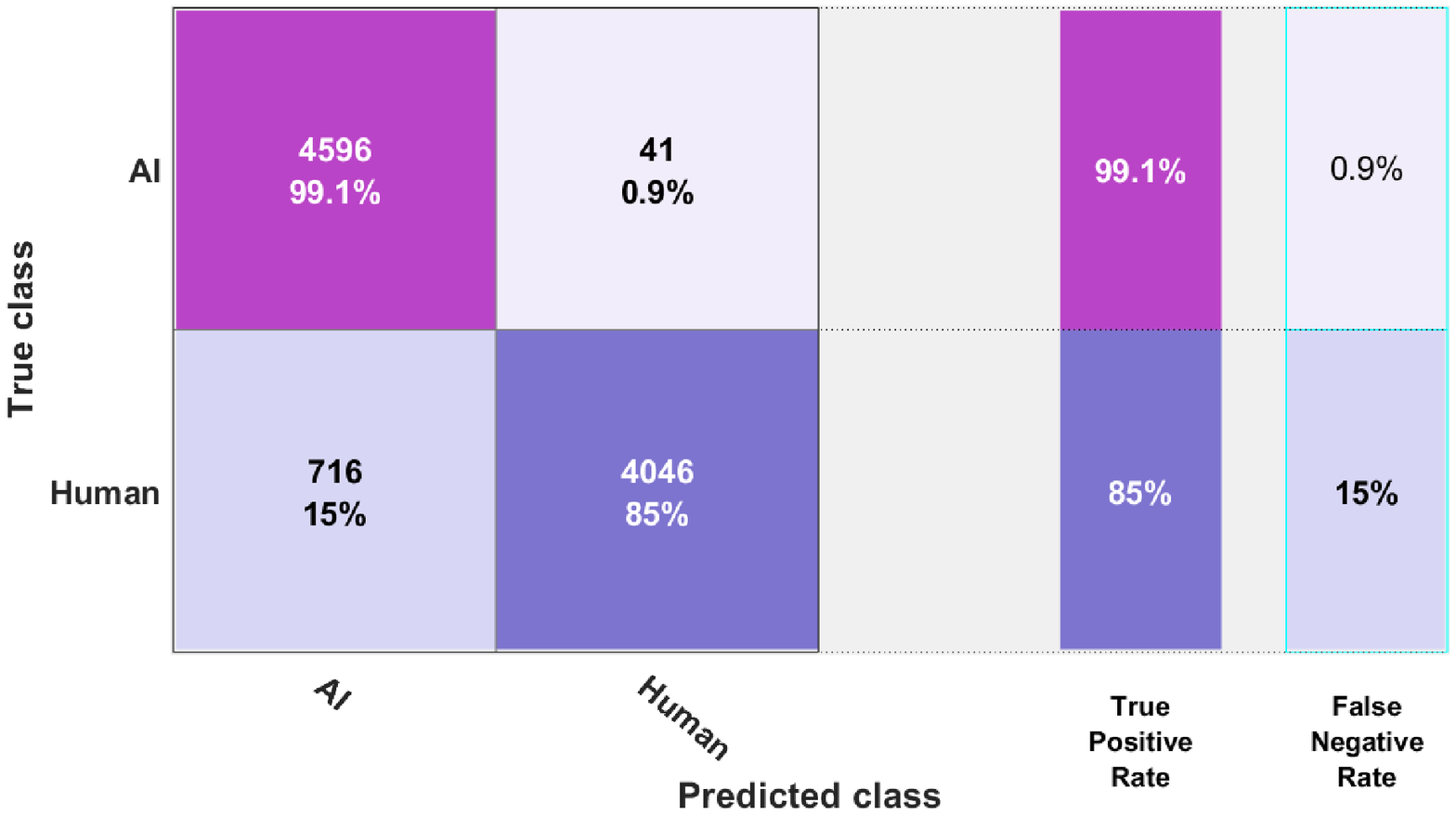}}
\caption{Confusion matrix for binary classification on test data set}
\label{fig7}
\end{figure}

\begin{figure}[ht!]
\centerline{\includegraphics[width=0.52\textwidth]{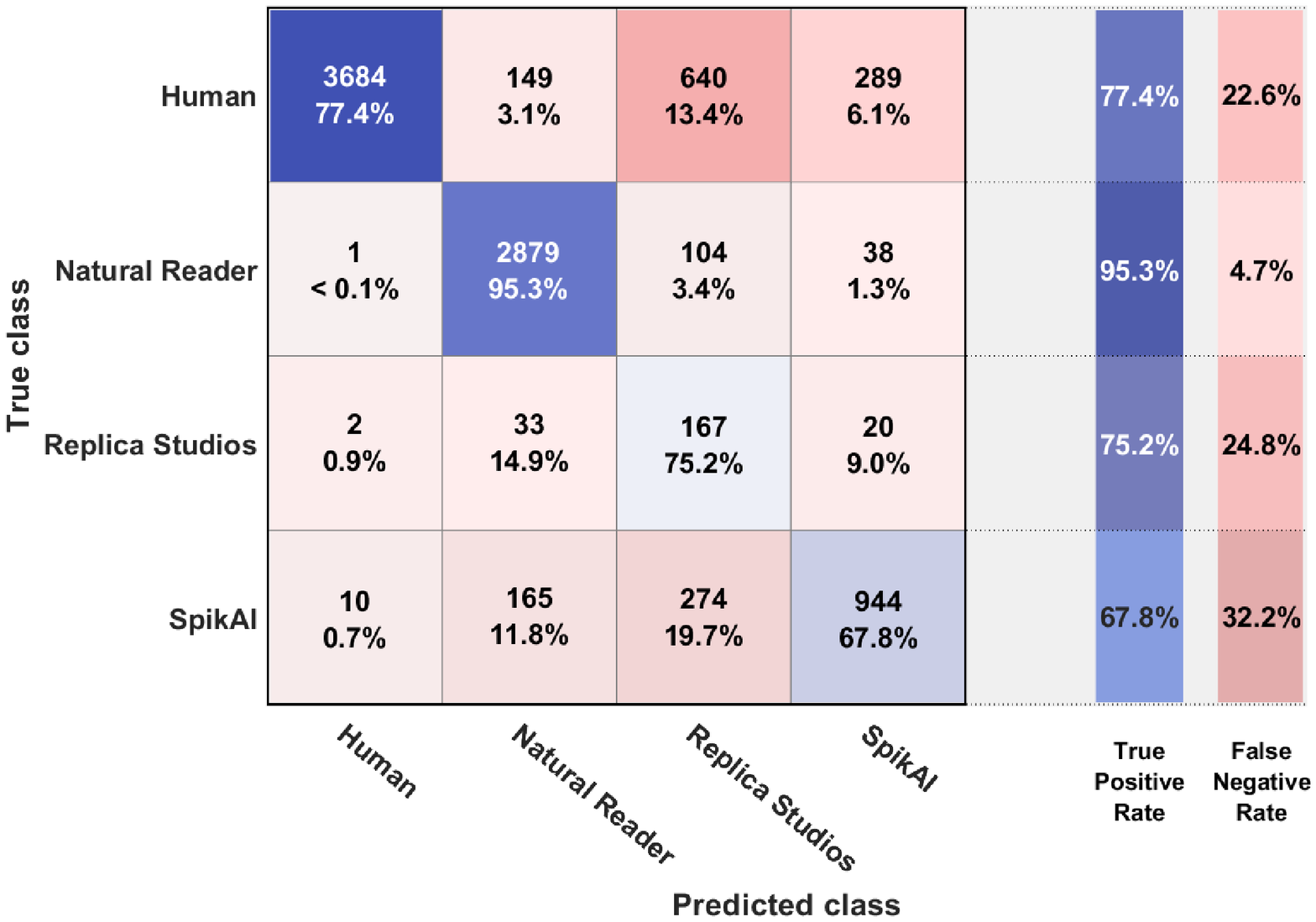}}
\caption{Confusion matrix for multi-class classification on test data set}
\label{fig8}
\end{figure}

After choosing our classifier and training the model, we tested our test data set over the trained model for the prediction. The test data consist of $9399$ random samples from all four types of speeches. Prediction is performed for both binary class classification and multi-class classification.  Figure \ref{fig7} and \ref{fig8} represent the confusion matrix for our final result of predicted data on binary classification and multi-class classification respectively.

The accuracy achieved on the prediction of test data on binary classification, which was our main motto, is 91.94\% with a miss classification rate of 8.06\%. For multi-class classification, we achieve an accuracy of 81.64\%, with a miss classification rate of 18.36\%.\

Also, we have calculated the F1 score for recognition of synthetic speech. The F1 score is given by :
\begin{equation}
F1\ Score = 2 \cdot \frac{precision\ \cdot\ recall}{precision + recall}
\end{equation}

where precision and recall are respectively given by:

\begin{equation}
precision = \frac{true\ positive }{true\ positive + false\ positive}
\end{equation}

\begin{equation}
recall = \frac{true\ positive }{true\ positive + false\ negative}
\end{equation}
F1 score values lies in the range of $0$ to $1$. Score near to $1$ represent high accuracy and and near to $0$ represent low accuracy. We achieve the F1 score of $0.923$ in recognising the synthetic speech.

\begin{table*}[ht]
\caption{Binary Class: F1 Score for Individual Features and few  of the combined Features over test data for different Machine Learning Models.}
\centering
{%
\begin{tabular}{|c|c|c|c|c|c|}
\hline
                          & \multicolumn{2}{c|}{\textbf{Multi Class}} & \multicolumn{3}{c|}{\textbf{Binary Class}}       \\ \hline
\multirow{2}{*}{\textbf{Various Models}} &
  \multicolumn{2}{c|}{All Features} &
  \multicolumn{3}{c|}{All Features} \\ \cline{2-6} 
 &
  \multicolumn{1}{l|}{{Training}} &
  \multicolumn{1}{l|}{{Testing}} &
  \multicolumn{1}{l|}{{Training}} &
  \multicolumn{1}{l|}{{Testing}} &
  \multicolumn{1}{l|}{{F1 Score}} \\ \hline
Fine Tree                 & 80.5               & 79.3                 & 86.5          & 82.9           & 0.8412          \\ \hline
Linear Discriminiant      & 69                 & 59.91                & 83.7          & 70.73          & 0.7681          \\ \hline
Quadratic Discriminant    & 67.7               & 54.92                & 76.3          & 85.51          & 0.8583          \\ \hline
Gaussian Naive Bayes      & 63                 & 57.03                & 61.5          & 88.23          & 0.8731          \\ \hline
Linear SVM                & 67.3               & 57.32                & 84.5          & 69.27          & 0.7595          \\ \hline
Quadratic SVM             & 70.8               & 71.73                & 84.7          & 78.59          & 0.8138          \\ \hline
Weighted KNN              & 70.5               & 65.3                 & 85.7          & 76.07          & 0.7998          \\ \hline
Boosted Trees Ensemble    & \textbf{83}        & 79.93                & 91.7          & 87.65          & 0.888           \\ \hline
Bagged Trees Ensemble     & 82.8               & \textbf{81.56}       & \textbf{93.7} & 86.5           & 0.8795          \\ \hline
RUSBoosted Trees Ensemble & 78.3               & 73.93                & \textbf{93.2} & \textbf{92.04} & \textbf{0.9246} \\ \hline
Logistic Regression       & --                 & --                   & 84.7          & 77.63          & 0.8125          \\ \hline
\end{tabular}%
} \vspace{2mm}
\label{tab:All_Features_Experiment}
\end{table*}

\subsection{Deep Learning Models Based Experiments}

Using CRNN32 as our deep learning model, for multi-class classification, we obtained an accuracy of 96.2\% accuracy on training data,  95.7\% accuracy on validation data, and  96.9\% accuracy on test data. Similarly, using same model for binary class classification, we obtained an accuracy of 97.9\% accuracy on training data, 97.1\% accuracy on validation data, and 98.1\% accuracy for test data. \

We chose the spectrogram image as a mode of our input in CRNN32 model, as spectrograms represent the characteristic features of audio, sounds, voices, speeches, etc. in form of patterns in an image. CRNN32 model employs CNN and BiLSTM in the construction of it's layers. The reason to choose the CNN in our model is due to the fact that CNNs are quite robust and exploit convolutional operation in extraction of features from these spectrogram image patterns.\


 BiLSTM are the best RNN models when the learning problem is sequential and there are dependencies among the past and future data. Hany has shown that long range temporal dependencies are induced into the synthetic audio during the process of synthesis \cite{b1}. Hence, BiLSTMs was chosen to capture these temporal dependencies.  
 

\begin{figure}[ht]
\centerline{\includegraphics[width=0.5\textwidth]{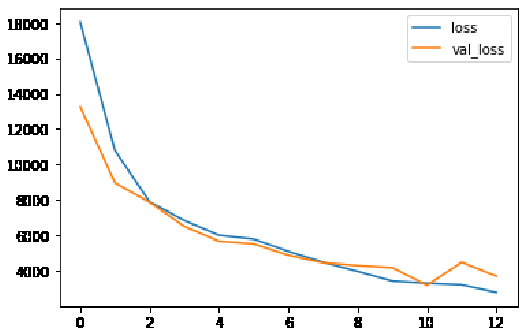}}
\caption{Plot for loss over training and validation data over number of epochs}
\label{binary_graph}
\end{figure}

\begin{figure}[ht]
\centerline{\includegraphics[width=0.5\textwidth]{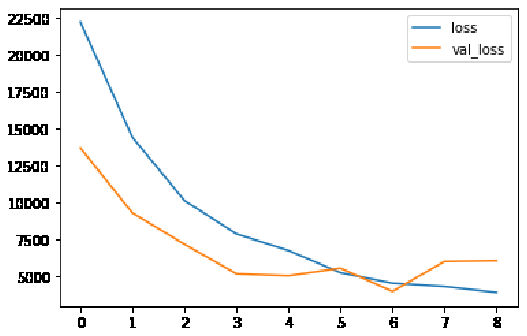}}
\caption{Plot for loss over training and validation data over number of epochs}
\label{fig9}
\end{figure}

\begin{figure}[ht]
\centerline{\includegraphics[width=0.52\textwidth]{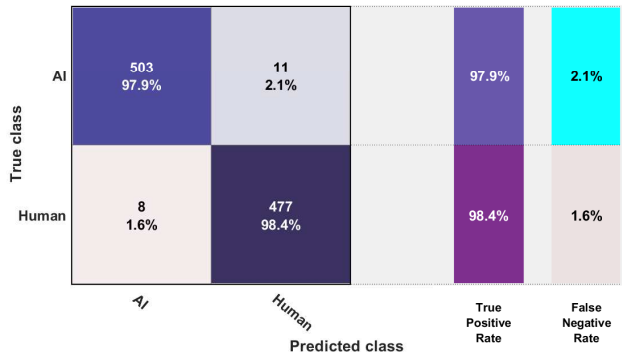}}
\caption{Confusion matrix for binary-class classification on test data set using CNN+RNN}
\label{fig10}
\end{figure}

\begin{figure}[ht]
\centerline{\includegraphics[width=0.52\textwidth]{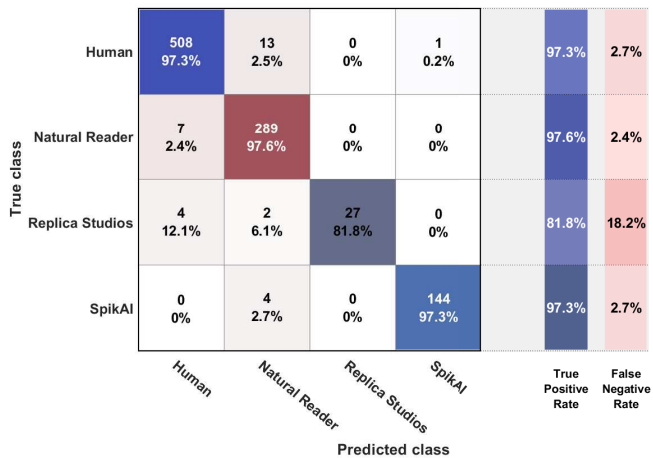}}
\caption{Confusion matrix for multi-class classification on test data set using CNN+RNN}
\label{fig11}
\end{figure}

\section{Conclusion}

From both parts of our experiments, in which we used Machine Learning and Deep Learning based models, we saw that our CRNN32 model of deep learning outperforms our machine learning model based approach and gave the better accuracy. But given the time of training which was less for our machine learning approach, we can see the accuracy achieved by our handcrafted features in machine learning for binary class classification is giving at par results. We believe that given the case scenarios, both our techniques can act as a good agent to detect the AI synthesized speech depending upon the application. \

The future work for this problem may include study and integration of other discriminatory features to improve upon the accuracy and decrease the miss classification rate. Also, the scalability of the proposed model can be validated by testing with more massive datasets. More variants of experiment scenarios like classification based on gender, age, and accent can be done. Due to evolution of amazing synthetic speech synthesizers, we are observing increased usage of synthetic speeches in native languages too. Hence, as a future direction, we plan to extend this research for identifying  synthetic speeches in other native languages.

\clearpage




%






\newpage

\end{document}